%% file: main.tex
\documentclass[10pt,twocolumn,letterpaper]{article}

\usepackage[pagenumbers]{cvpr} 
\usepackage[framemethod=TikZ]{mdframed}
\usepackage{makecell}

\usepackage{colortbl}
\usepackage{multirow}

\definecolor{cvprblue}{rgb}{0.21,0.49,0.74}
\usepackage[pagebackref,breaklinks,colorlinks,allcolors=cvprblue]{hyperref}

\def\paperID{8800} 
\def\confName{CVPR}
\def\confYear{2025}

\title{Octopus: Alleviating Hallucination via Dynamic Contrastive Decoding}

\author{Wei Suo$^{1}$\footnotemark[1] , Lijun Zhang$^{1}$\footnotemark[1], Mengyang Sun$^{2}$, 
Lin Yuanbo Wu$^{3}$, Peng Wang$^{1}$\footnotemark[2] , Yanning Zhang$^{1}$
\\
$^1$School of Computer Science and Ningbo Institute, Northwestern Polytechnical University,China.\\
$^2$School of Cybersecurity, Northwestern Polytechnical University, China.\\
$^3$Computer Science, Swansea University.\\
{\tt\small \{suowei1994,lijunzhang,sunmenmian\}@mail.nwpu.edu.cn
\tt\small \{peng.wang,ynzhang\}@nwpu.edu.cn
}}

\begin{document}
\maketitle
\renewcommand\thefootnote{\fnsymbol{footnote}}
\footnotetext[1]{These authors contributed equally to this work.} 
\footnotetext[2]{Corresponding authors.} 

\input{sec/0_abstract}    
\begin{tikzpicture}[remember picture,overlay,shift={(current page.north west)}]
\node[anchor=north west,xshift=1.8cm,yshift=-3.4cm]{\scalebox{-1}[1]{\includegraphics[width=1.0cm]{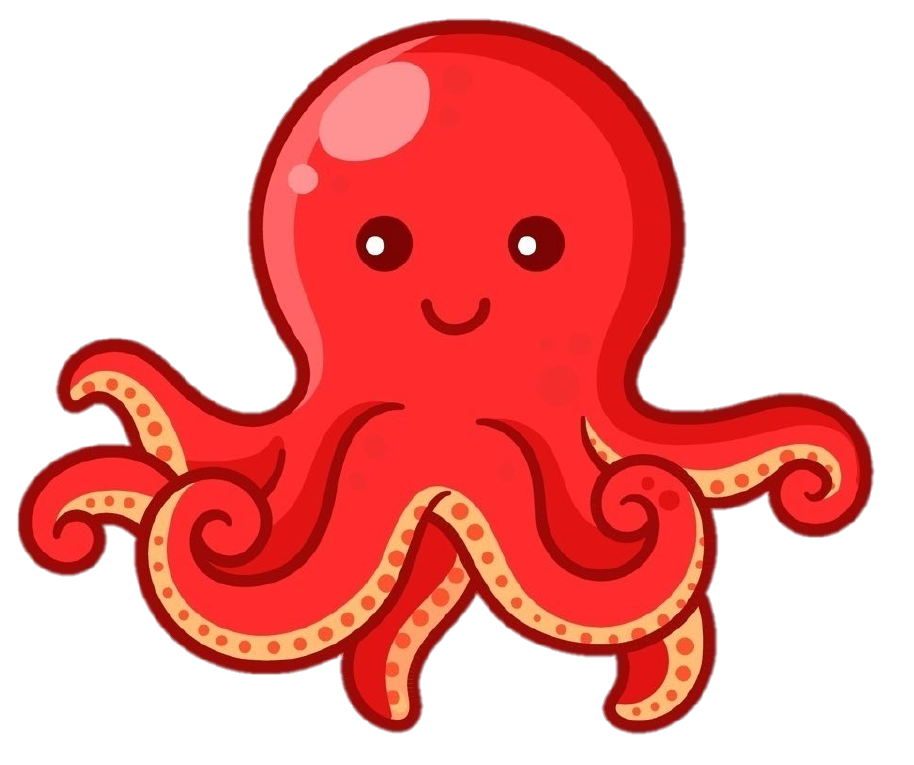}}};
\end{tikzpicture}
\input{sec/1_intro}
\input{sec/2_related}

\input{sec/3_preliminary}

\input{sec/4_method}

\input{sec/5_exper}
\input{sec/6_conclusion}
\clearpage
{
    \small
    \bibliographystyle{ieeenat_fullname}
    \bibliography{main}
}
\appendix


\end{document}

%% file: sec/0_abstract.tex
\begin{abstract}

Large Vision-Language Models (LVLMs) have obtained impressive performance in visual content understanding and multi-modal reasoning. Unfortunately, these large models suffer from serious hallucination problems and tend to generate fabricated responses. Recently, several Contrastive Decoding (CD) strategies have been proposed to alleviate hallucination by introducing disturbed inputs. Although great progress has been made, these CD strategies mostly apply a one-size-fits-all approach for all input conditions. In this paper, we revisit this process through extensive experiments. Related results show that hallucination causes are hybrid and each generative step faces a unique hallucination challenge. Leveraging these meaningful insights, we introduce a simple yet effective Octopus-like framework that enables the model to adaptively identify hallucination types and create a dynamic CD workflow. Our Octopus framework not only outperforms existing methods across four benchmarks but also demonstrates excellent deployability and expansibility. Code is available at \href{https://github.com/LijunZhang01/Octopus}{https://github.com/LijunZhang01/Octopus}.

\end{abstract}

%% file: sec/1_intro.tex
\section{Introduction}
\label{sec:intro}

Large Vision-Language Models (LVLMs) have achieved significant success over the past few years~\cite{bai2023qwen,10.5555/3666122.3668264,liu2024improved,liu2024visual,zhu2023minigpt}.
They have facilitated various Vision-and-Language (VL) tasks~\cite{suo2023s3c,li2019transferable,suo2022rethinking,hossain2019comprehensive} by adapting to different input instructions.
However, LVLMs are facing a grand challenge: they often fail to accurately capture the visual content and
tend to generate fabricated responses (\emph{e.g.,} imaginary objects, 
incorrect attributes and inexistent relationship), which is known as \textit{hallucination}~\cite{gunjal2024detecting,liu2023mitigating}.  
The hallucination issue seriously affects user trust and confidence, especially in applications that require trustworthy outcomes such as medical reports~\cite{fan2020inf,xie2018fusing} and automatic driving ~\cite{dai2019hybridnet}. 

\begin{figure}[t]
\centering
\includegraphics[width=0.48\textwidth]{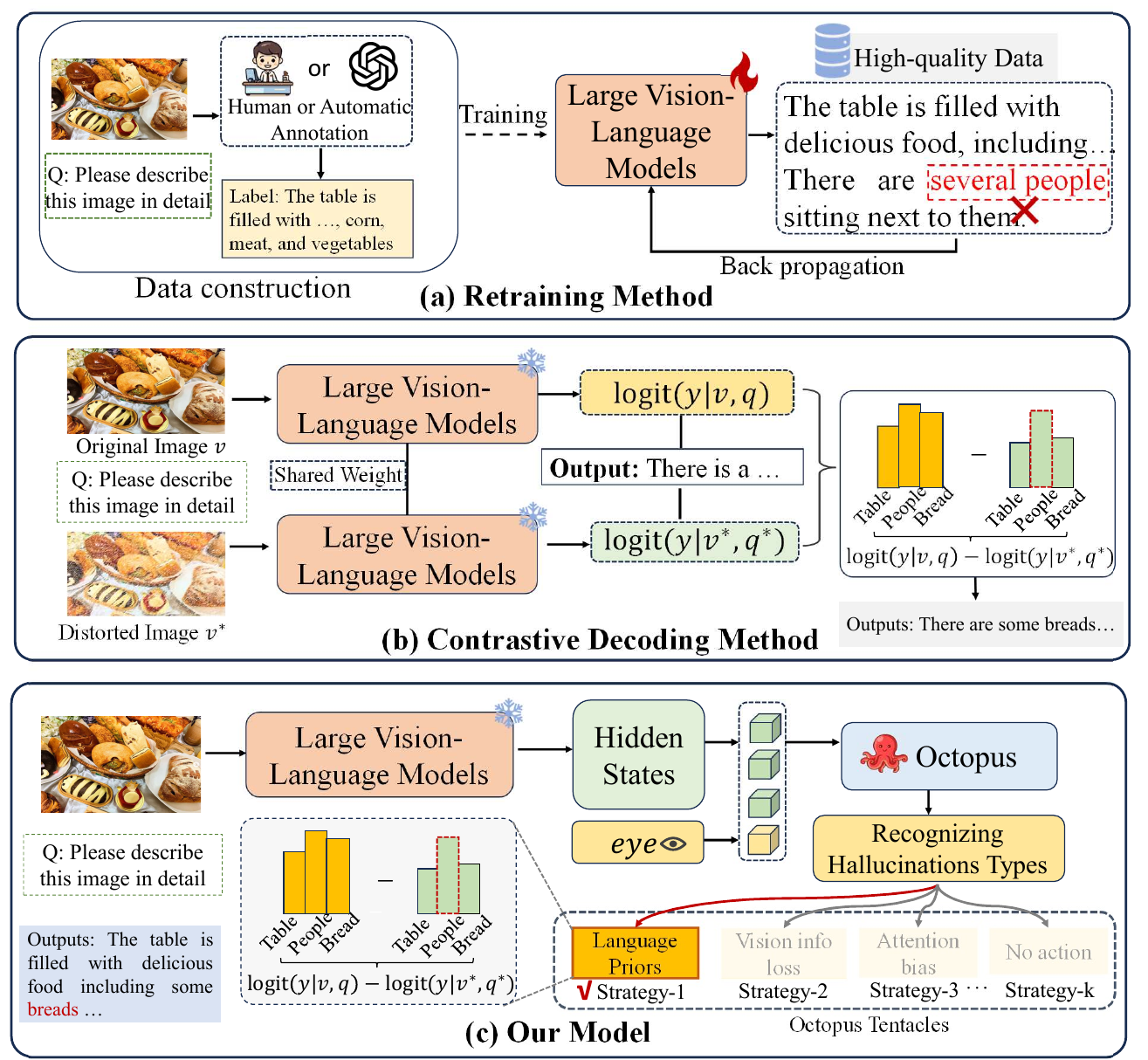}
\caption{Paradigm comparison of different hallucination alleviation methods. (a) Retraining method. Constructing high-quality data to retrain these LVLMs. (b) {Contrastive Decoding.} Comparing the output distributions from the original and distorted inputs. (c) {Octopus.} Our method focuses on dynamically selecting suitable strategies to reduce hallucinations caused by various factors.}
\label{fig:fig1}
\end{figure}

To alleviate the hallucination issue, existing approaches can be roughly categorized into two research lines. As shown in Fig.~\ref{fig:fig1} (a), the first paradigm relies on constructing high-quality instruction tuning data and re-training the models to suppress  hallucinations~\cite{gunjal2024detecting,liu2023mitigating,lu2024evaluation,wang2024vigc}. However, such a strategy requires a well-designed data construction process with complex verification and expensive costs. In addition, additional training is strictly prohibited for deployed models. 
In contrast, as shown in Fig.~\ref{fig:fig1} (b),  Contrastive Decoding (CD) methods ~\cite{leng2024mitigating,favero2024multi,woo2024don,manevich2024mitigating} aim to
contrast the logit scores at each generative step from the original inputs (\emph{i.e.,} image input $v$ and corresponding text $q$) with those derived from the modified inputs (\emph{i.e.,} $v^{*}$ and $q^{*}$). As a post-hoc corrective mechanism, such a methodology can effectively eliminate hallucination without complex training. 
In practice, these CD methods 
focus on designing a new pair of $v^{*}$ and $q^{*}$ to alleviate hallucination problems. 
For example,~\cite{leng2024mitigating} utilizes Gaussian noise to mask the original image and overcome language priors.~\cite{woo2024don} employs enhanced visual input to mitigate attention bias.

Despite their strong performance, these CD methods mostly apply the same disturbed manner for all samples and generative steps. 
This motivates us to question: 
1) Is a single CD strategy suitable for different samples? 2) Do all generative steps (\emph{i.e.,} predicted tokens) experience the same type of hallucination? To answer the two questions, we conduct exploratory experiments to understand the causes of hallucinations at both sample and token levels. Specifically, we use a group of CD methods as off-the-shelf diagnostic tools to investigate the pattern of hallucinations. Our analysis reveals that 
each CD method is only effective on specific hallucinated samples, and using a single CD strategy would lead to sub-optimal results.
Meanwhile, we thoroughly investigate the process of hallucination emergence at the token level. Through an enumeration method~\cite{beckwith1966process} and qualitative analysis, we find that fabricated responses are hybrid and each generative step faces a unique hallucination issue. 

The above results indicate that a one-size-fits-all approach struggles to correct different types of hallucinations effectively. 
Thus a natural idea is to combine multiple CD methods to reduce hallucinations from various sources. However, without well-defined labels, identifying the optimal strategy for different input samples is challenging. Additionally, token generation is sequentially dependent and involves a vast solution space, making it difficult to choose the best CD approach at each generative step.
To tackle the above problems, as shown in Fig.\ref{fig:fig1} (c), we introduce a simple yet effective framework, called
\textbf{Octopus}. 
Different from previous works, our method focuses on guiding the model to dynamically organize the contrastive decoding workflow and selecting the appropriate CD strategy based on different inputs. In particular, we first build a transformer-based block and a learnable token to adaptively recognize the type of hallucination, similar to the Octopus's eyes.
According to different decisions, each CD strategy is regarded as a ``tentacle'' to perform a specific contrastive operation. 
Finally, leveraging Direct Preference Optimization (DPO)~\cite{rafailov2024direct} or reinforcement learning~\cite{wang2012monte, schulman2017proximal}, Octopus can be easily optimized. 
Benefiting from the above designs, the proposed method not only effectively reduces hallucinated content, but also scales well for deployments due to avoiding retraining weights of LVLMs. More importantly, as a general framework, subsequent CD methods can be seamlessly integrated without additional adjustments.
In summary, we make the following contributions:

1) Our work reveals that the mechanism of hallucination occurrence is a complex hybrid and different samples (or tokens) suffer from various forms of hallucination challenges. 

2) We develop a new framework Octopus that can adaptively recognize the types of hallucination and build a dynamically contrastive decoding workflow to correct fabricated content. 

3) Octopus achieves state-of-the-art performance across four benchmarks for both generative and discriminative tasks, while also demonstrating excellent deployability and expansibility.

%% file: sec/2_related.tex
\section{Related work}
\label{sec:related}
\subsection{Large Visual-Language Models}
Large Visual-Language Models (LVLMs)~\cite{bai2023qwen,10.5555/3666122.3668264,liu2024improved,liu2024visual,zhu2023minigpt} usually
consist of three key components:
a visual encoder like CLIP~\cite{radford2021learning}, a Large Language Model (LLM) such as LLAMA~\cite{touvron2023llama} and a cross-modal alignment module that connects the visual encoder's output to the LLM. 
LVLMs have obtained impressive performance in visual content understanding and multi-modal reasoning such as image captioning~\cite{hossain2019comprehensive,zhu2023minigpt}, referring expression comprehension~\cite{yue2024sc,liu2023grounding}, 
and visual question answering ~\cite{dang2024sadl,wang2024surgical}. However, LVLMs still face significant hallucination issues, which seriously limit their practical usability.

\subsection{Hallucination in LVLMs}
To alleviate the hallucinations in LVLMs, both data-driven retraining and Contrastive Decoding (CD) methods have been proposed.
Data-driven methods aim to enhance data quality to reduce the hallucinations~\cite{gunjal2024detecting,liu2023mitigating,lu2024evaluation,wang2024vigc}. For example, some works introduce negative data~\cite{liu2023mitigating} and counterfactual data~\cite{yu2024hallucidoctor} to mitigate hallucination issues. 
~\cite{wang2024mitigating} cleans the dataset to minimize noise and errors. ~\cite{yu2024rlhf,wen2024policy} annotates a high-quality hallucination dataset to suppress the occurrence of hallucinations by fine-tuning.
In contrast, CD methods tackle hallucinations by comparing output distributions from original and distorted input without altering the model's weights. 
For instance,~\cite{leng2024mitigating} alleviates hallucinations by counteracting language priors, while~\cite{favero2024multi} tackles them through refined visual prompts.~\cite{chen2024halc} leverages multi-perspective visual cues to correct false information.
Different from the above methods, our work focuses on adaptively selecting the most suitable CD strategies to alleviate different hallucination issues.

%% file: sec/3_preliminary.tex
\section{Preliminary}


\label{sec:preliminary}


\subsection{Large Vision Language Models}
\label{subsec:preliminary_cd}
Given a Large Vision-Language Model (LVLM) with parameters  $\theta$, the model can effectively perform various multi-modal tasks using a visual input $v$ and a textual instruction $q$. Specifically, at each generative step $t$, the auto-regressive LVLM conducts the following calculations:
\begin{equation}
\label{eq:1}
     \ell_t = {\rm{log}}\ p(y_t|v,q,y_{<t};\theta),
\end{equation}
\begin{equation}
     y_t \sim \rm{Softmax}(\ell_t),
\end{equation}
where $\ell_t$ is the logit score for the next token $y_t$, while $y_{<t}$ represents the response generated before time step $t$. After extensive training, well-designed LVLMs demonstrate impressive understanding ability across a wide range of multi-modal tasks. However, these models suffer from serious hallucination issues~\cite{gunjal2024detecting,liu2023mitigating,lu2024evaluation,wang2024vigc}. 
They often produce inaccurate answers or fabricated descriptions that may not align with the visual input.



\subsection{Contrastive Decoding}
\label{sec:sec3.2}
To mitigate cross-modal hallucinations, Contrastive Decoding (CD) offers a promising approach by contrasting output distributions between original and distorted inputs. In particular, CD methods first generate two output distributions: one from the original visual image $v$ and textual query $q$, and another from the perturbed inputs $v^*$ and $q^*$.
Then, by examining the difference between two distributions, a contrastive response $\ell_{cd}$ can be formulated as follows:
\begin{equation}
\begin{aligned}
\label{lcd}
\ell_{cd}=m{{\rm{log}}\ p(y_t|v,q,y_{<t};\theta)}- \\ n{{\rm{log}}\ p(y_t|v^*,q^*,y_{<t};\theta)},
\end{aligned}
\end{equation}
\begin{equation}
     y_t \sim \rm{Softmax}(\ell_{cd}),
\end{equation}
where $m$ and $n$ are hyperparameters, and 
$y_t$ represents the predicted token based on the contrastive decoding. 
Although CD methods are effective in mitigating hallucinations, they generally apply the same disturbed manner to all samples and generative steps. 
In this paper, we rethink the above operations with extensive experiments. 
Considering that existing CD methods are often tailored to specific types of hallucinations, we select three well-received CD methods as diagnostic tools to further investigate the mechanisms underlying hallucination occurrence. We will introduce them in detail.


Strategy-1: VCD~\cite{leng2024mitigating}. VCD focuses on \textbf{\textit{overcoming language priors}}. 
It employs a Gaussian noise mask to generate the distorted visual input $v^*$, while the query text $q$ is unchanged. 

Strategy-2: M3ID~\cite{favero2024multi}. M3ID relieves hallucinations by \textbf{\textit{reducing visual information loss}}.
It masks the query text to build $q^{*}$ and independently supplies the visual input $v$ into the LVLMs as the distorted input. 

Strategy-3: AVISC~\cite{woo2024don}. AVISC alleviates hallucinations by 
\textbf{\textit{minimize attention bias}}.
It constructs a new visual input $v^*$ using blind tokens that do not contain information relevant to the query.

These three strategies correspond to different causes of hallucinations: \textbf{\textit{language priors, vision information loss and attention bias}}. 
Next, we conduct related experiments on the popular VL model LLaVA-1.5-7B~\cite{liu2024improved} to explore the two key questions: 1) Is a single CD strategy effective for all multi-modal samples? 2) Does each time step experience the same cause of hallucination?

\begin{figure}[t]
\centering
\includegraphics[width=0.48\textwidth]{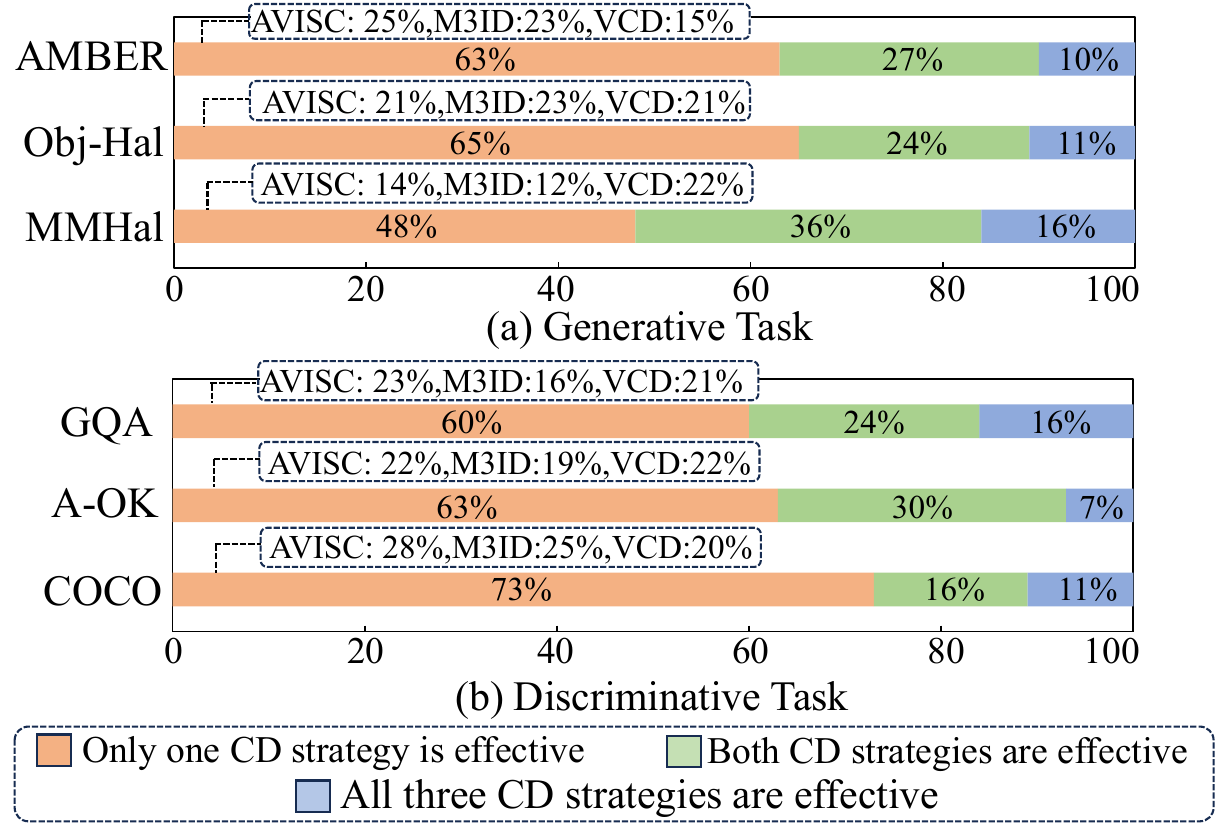}
\caption{The proportion of effective samples using different CD methods for (a) Generative Task and (b) Discriminative Task. We observe that each CD strategy can only address part of the samples.}
\label{fig:fig2}
\end{figure}
\label{subsec:token}

\subsection{Sample-Level Hallucination}
\label{subsec:sample}
In this section, we conduct two kinds of experiments (\emph{i.e.,} generative and discriminative tasks) to answer the first question.
For generative task, we establish experiments on the three widely used datasets: 
AMBER~\cite{wang2023llm}, Object-HalBench~\cite{rohrbach2018object} 
with the language prompt ``Please describe this image in detail'', as well as MMHalbench~\cite{sun2023aligning} with the original instructions as the prompts. 
Following~\cite{wang2024mdpo}, both the AMBER and Object-HalBench use CHAIR score~\cite{rohrbach2018object} as metrics to evaluate the degree of hallucination, while the MMHalbench uses the GPT-4V score~\cite{sun2023aligning}. In addition, we use these three strategies in Sec.~\ref{sec:sec3.2} to interfere with the output distributions of LLaVA for each sample, respectively. By utilizing the above metrics, this strategy is identified as effective when it attains better performance compared to the original LLaVA output. As shown in Fig.~\ref{fig:fig2} (a), 
we report the corresponding percentages, where the orange, green and blue denote ``Only One CD strategy is effective'', ``Both CD strategies are effective'' and ``All three CD strategies are effective'', respectively. By comparing these results, we observe that a large number of samples ($\sim$60\%) can only be addressed by certain specific CD strategies and their overlap is relatively small ($\sim$10\%).

For discriminative task, we conduct similar experiments on the three POPE datasets~\cite{li2023evaluating} (\emph{i.e.,} GQA~\cite{hudson2019gqa}, A-OKVQA~\cite{schwenk2022okvqa} and COCO~\cite{lin2014microsoft}) with the language template ``$Q$ + Please answer this question in one word'', where $Q$ represents the textual question. Different from the generative task, we directly apply strategies 1-3 to each question and count the number of samples whose answers are corrected. As shown in Fig.~\ref{fig:fig2} (b), we report the corresponding percentages and observe that each CD strategy addresses a subset of the samples, and only $\sim$10\% of cases are effective across all three methods. Based on the above results, we conclude that \textbf{\textit{each CD method is only effective on specific hallucinated samples, and using a single strategy for all cases would inevitably lead to sub-optimal results.}} 


\subsection{Token-Level Hallucination}
\begin{figure}[t]
\centering
\includegraphics[width=0.48\textwidth]{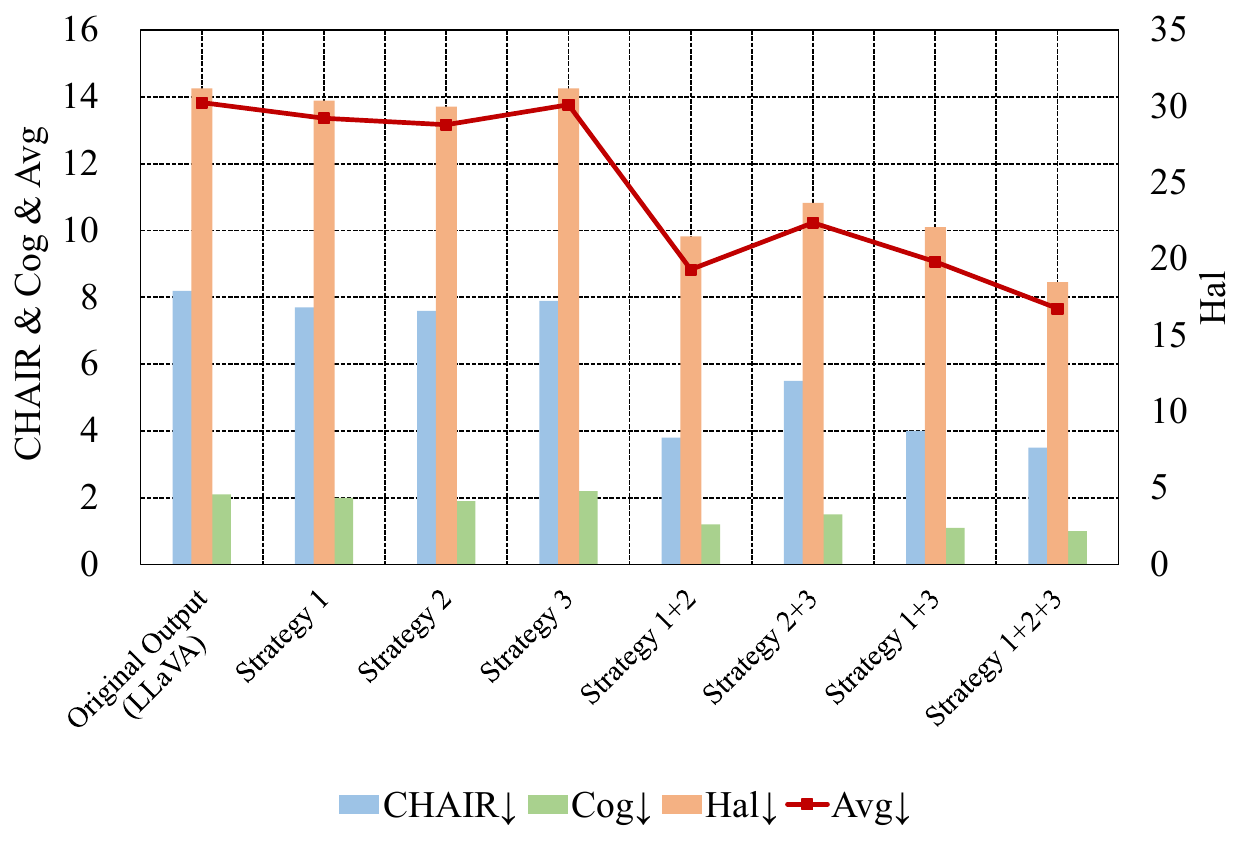}
\caption{Token-level hallucination 
quantitative evaluation. We enumerate different CD strategies at each time step. The results show that using multiple CD strategies obtains better performance.}
\label{fig:fig2_add}
\end{figure}

The above sample-level analyses demonstrate that each sample needs to adopt a specific CD strategy. 
In this section, we focus on a more fine-grained scenario: whether each time step in the generative process suffers from the same hallucination causes. 

To counteract this, we construct experiments on the AMBER~\cite{wang2023llm} dataset with three metrics: CHAIR~\cite{rohrbach2018object}, Cog~\cite{wang2023llm} and Hal~\cite{wang2023llm}. Because these CD strategies in Sec.~\ref{sec:sec3.2} can be regarded as three types of diagnostic tools, we apply the enumeration method~\cite{beckwith1966process} to evaluate the hallucinatory causes in the generative process. 
Moreover, we are aware that even though there are just three CD candidates, the combination space is still enormous due to the lengthy outputs. To reduce the number of combinations, the enumerating space would only consider the first three hallucinated nouns in each description. As shown in Fig.\ref{fig:fig2_add}, we use ``strategy-1'', ``strategy-2'' and ``strategy-3'' to denote the hallucination mitigation strategies introduced by Sec.~\ref{sec:sec3.2} (\emph{i.e.,} VCD~\cite{leng2024mitigating}, M3ID~\cite{favero2024multi} and AVISC~\cite{woo2024don}). Meanwhile, we exhibit the best scores from these combinations. Take “strategy 1+3” as an example, each of the three tokens has two selectable hallucination elimination strategies (\emph{i.e.,} strategy-1 and strategy-3), thus there are a total of 6 combinations. For simplicity, we only report the best results among these combinations. By comparing these scores, we find that leveraging multiple CD strategies can better suppress hallucinations. 
\begin{figure}[t]
\centering
\includegraphics[width=0.48\textwidth]{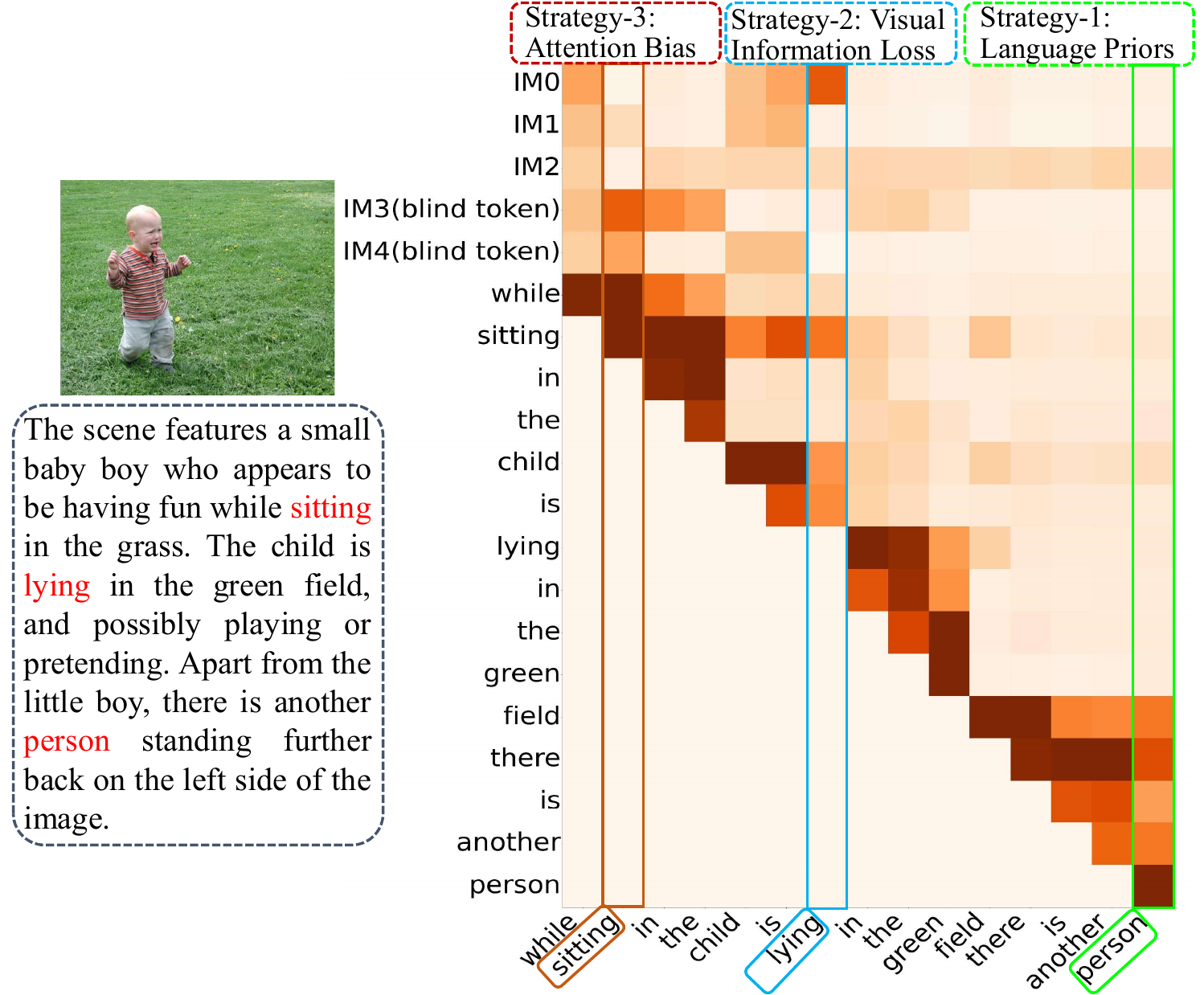}
\caption{Token-level hallucination qualitative analysis. For simplicity, we only present the attention map across the top 5 visual tokens and corresponding keywords. The results show that hallucination causes are hybrid in a sample.}
\label{fig:dingxin_attention_gen}
\end{figure}

To investigate the nature of hallucination occurrence, we also conduct a qualitative analysis to examin the attention distribution for each predicted token. As shown in Fig.~\ref{fig:dingxin_attention_gen}, 
it can be found that hallucinated words include ``sitting, lying, and person'' in this sentence, where each token corresponds to different causes of hallucination. For example, the ``sitting'' focused on the visual blind token ``IM3'', indicating that the current step is affected by attention bias~\cite{woo2024don}. The occurrence of ``lying'' is primarily due to insufficient attention to visual information~\cite{favero2024multi}. While the ``person'' concentrates solely on language tokens, suggesting that it is influenced by language priors~\cite{leng2024mitigating}. 
Therefore, we conclude that \textbf{\textit{the hallucination causes are hybrid and each generative step faces different forms of challenge.}} 

\begin{figure*}[t]
  \centering
    \includegraphics[width=1\textwidth]{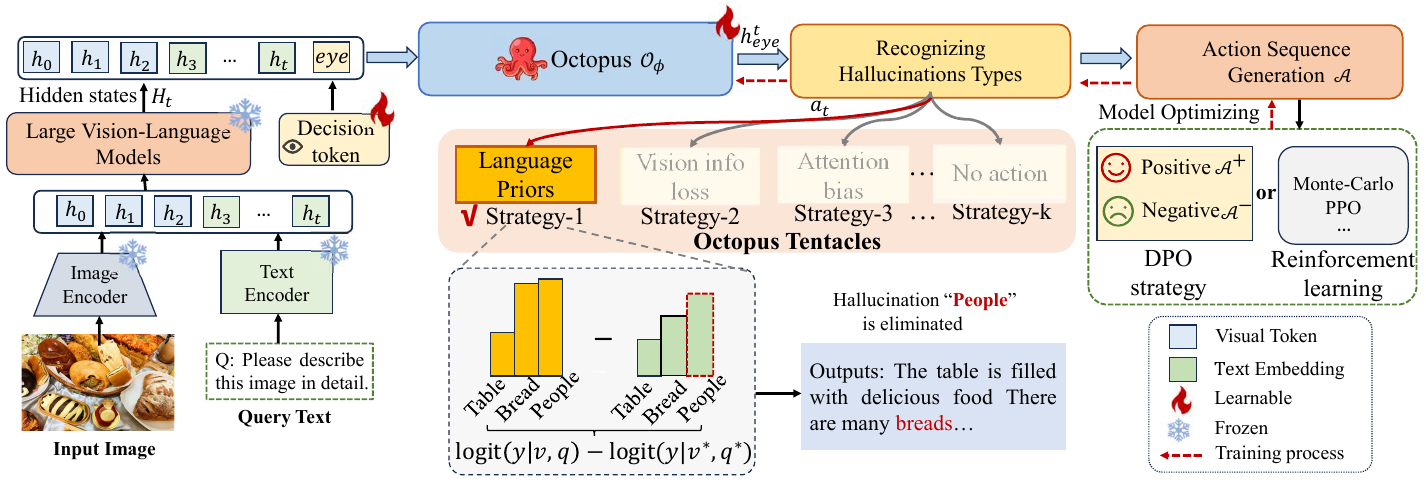}  
  \caption{\textbf{Overview of our method.} 
  Our Octopus framework consists of two key components: the decision token $eye$ and its tentacles. Specifically, we first utilize the ``$eye$'' to identify the types of hallucinations, and then these ``tentacles'' are applied to address specific hallucination issues at each generative step.
   Finally, our model would be optimized by DPO or other reinforcement learning methods.}
  \label{fig:fig3}
\end{figure*}

\subsection{Discussion}
\label{subsec:discussion}
Based on the above experiments, it can be found that various hallucination factors collectively lead to falsity outputs across the sample and token levels. Therefore, a natural idea is to combine these off-the-shelf CD strategies as corrective approaches and tackle different types of hallucinations.  
However, due to lacking pre-defined labels, it is difficult to select the most suitable strategy for each sample. Meanwhile, since the textual generation process is sequentially dependent, the choice of hallucination elimination strategy for each token would be influenced by the previous selection. Considering the vast solution space, deciding which CD strategy to use is challenging at each generative step.

%% file: sec/4_method.tex
\section{Our Method}


To alleviate the above challenges, we propose a simple yet effective framework Octopus, which can adaptively select the proper CD strategy and organize the contrastive decoding workflow based on different input conditions. 
As shown in Fig.\ref{fig:fig3}, instead of relying on just one technology, our method focuses on leveraging an Octopus-like structure to detect hallucination types and perform corresponding contrastive actions. Considering that the discriminative task can be viewed as a generation sequence with a response length of 1, we will take the more complex generative task as an example to introduce our method.

\subsection{Model Structure}
\label{Octopus}
Given a vision input $v$ and a textual instruction $q$ (\emph{e.g.,} ``Please describe this image in detail''), we utilize the Octopus's eye to recognize the type of hallucination at each generative step. Then these Octopus's tentacles are used to perform specific strategies by contrastive decoding. 



Specifically, we first construct a vanilla transformer-based block $\mathcal{O}_{\phi}$, where $\phi$ denotes the parameters of the transformer structure~\cite{vaswani2017attention}. 
Based on Eq.\ref{eq:1}, it can be found that each time step $y_t$ would be influenced by $v$, $q$, and $y_{<t}$ together. 
Therefore, these hidden states from LVLMs (\emph{i.e.,} $v$, $q$ and $y_{<t}$) would be fed into the $\mathcal{O}_{\phi}$ with a decision token $eye\in\mathbb{R}^{d}$, where $d$ is the dimension of hidden state. For simplicity, we use $H_t=\{h_i\}_{i=1}^{t}$ to represent the sequence before $t$-th generation step, where $h_i\in\mathbb{R}^{d}$ is hidden state of each token. 
While the learnable token $eye$ can be regarded as ``Octopus's eye''. 
Formally, the above computations can be formulated as:
\begin{equation}
     [h_{eye}^t; H_t^{'}] = {\mathcal{O}_{\phi}}\ ({\rm{concat}} [eye ; H_t] + E_{pos}),
\end{equation}
where $h_{eye}^t$ and $H_t^{'}$ are 
corresponding outputs from $eye$ and $H_t$ sequence, respectively. While $E_{pos}$ and $\rm{concat}$ denote position embedding and concatenate operation. 
Benefiting from the self-attention mechanism~\cite{vaswani2017attention}, the $h_{eye}^{t}$ can adaptively aggregate the information from other hidden states. 

Then, a light and simple Multi-Layer Perceptron (MLP) is utilized to map the $h_{eye}^{t}$ into action vector $h_{act}^{t} \in\mathbb{R}^{k}$, where the $k$ is the number of candidate strategies. In this paper, we build four action spaces at each step including strategies 1-3 in the Sec.~\ref{sec:sec3.2} (\emph{i.e.,} VCD~\cite{leng2024mitigating}, M3ID~\cite{favero2024multi} and AVISC~\cite{woo2024don}) and a null action (\emph{i.e.,} no CD strategy is performed). Here, we use ``tentacles'' to represent these candidate CD actions.
For each $h_{act}^{t}$, the action vector $a_t$ is obtained by:
\begin{equation}
\label{fact}
     h_{act}^{t} = {\rm{MLP}}(h_{eye}^{t}),
\end{equation}
\begin{equation}
\label{a_t}
     a_t = {\rm{argmax}} ({\rm{Softmax}}(h_{act}^{t})),
\end{equation}
where $\rm argmax$ refers to the operation of selecting the index of the maximum value, while $\rm Softmax$ is the active function.
Based on the one-hot vector $a_{t}$, our Octopus can conveniently choose the corresponding CD strategies to implement. Finally, 
we would obtain a contrastive decoding workflow $\mathcal{A}=\{a_t\}_{t=1}^{N}$, where $N$ is the length of response. 

\subsection{Model Optimizing}
\label{subsec:dpo}

It can be noted that there is a non-differentiable operation in the above computations (\emph{i.e.,} Eq.\ref{a_t}), and the optimization process is also challenging due to the lack of explicit decision labels and serious curse of dimensionality~\cite{abel2016near}. Therefore, we introduce Direct Preference Optimization (DPO)~\cite{rafailov2024direct} to alleviate this problem. In fact, our Octopus can also be optimized using other Reinforcement Learning (RL) methods such as Monte-Carlo sampling~\cite{wang2012monte} or PPO~\cite{schulman2017proximal} (more discussion can be found in Sec.~\ref{duibi:rl}). Due to its simplicity and stability compared to other approaches~\cite{wang2012monte,schulman2017proximal}, we will only introduce the DPO optimization approach in this section.


\noindent
\textbf{Data Construction.}
The DPO method is designed to replace the typical RLHF procedure~\cite{ouyang2022training} and it can directly fit the human preference by building positive and negative samples. Inspired by this, we reformulate the above action choice process as a preference problem, which encourages our Octopus $\mathcal{O}_{\phi}$ to produce the action sequences that can effectively mitigate hallucinations. For constructing the positive workflow $\mathcal{A}^{+}$ and negative workflow $\mathcal{A}^{-}$, we generate 10 sequences for each sample by randomly selecting four actions at each generative step. Next, we divide them into $\mathcal{A}^{+}$ and $\mathcal{A}^{-}$ from these responses according to the CHAIR metric~\cite{rohrbach2018object}.  In practice, this metric is also flexible and can be adjusted for different fields. For the discrimination task, we separately use four ``tentacles'' to build $\mathcal{A}$, while the positive and negative samples are split by the answer's confidence scores. 
Finally, following~\cite{li2023silkie}, we use balanced positive and negative samples to construct the preference dataset. 

\noindent
\textbf{Training Process.}
To guide the policy model output preferred sequences $\mathcal{A}^{+}$, the traditional RL-based methods have to depend on a complex sampling process~\cite {wang2012monte} or an additional reward model~\cite{schulman2017proximal}. Conversely, the DPO replaces the reward process with a policy model and a reference model, which can straightforwardly
increase the likelihood of positive sequences. 
Therefore, given the above preference dataset $\mathcal{D}=\{\mathcal{A}^{+}, \mathcal{A}^{-}\}$, we apply the DPO to directly optimize our Octopus. 
In addition, previous works have proved that removing the reference model can obtain better or comparable performance than the original DPO~\cite{xu2023some,meng2024simpo}. Based on this, our optimization objective is defined as follows:
\begin{equation}
\begin{aligned}
     \max _{\mathcal{O}_{\phi}} \mathbb{E}_{\left(x,\mathcal{A}^{+} , \mathcal{A}^{-}\right) \sim \mathcal{D}} \log \sigma\left(\beta \log \mathcal{O}_{\phi}\left(\mathcal{A}^{+}\mid x\right)\right.\\\left.
     -\beta \log \mathcal{O}_{\phi}\left(\mathcal{A}^{-} \mid x\right)\right),
\end{aligned}
\label{eq:dpo}
\end{equation}
where $x=(v,q)$ is the input sequence, $\sigma$ denote sigmoid function. Following~\cite{li2023silkie}, we set the $\beta$ to 1. Based on the above training process, our Octopus can adaptively learn to construct a suitable workflow without human labeling. Moreover, note that our method would only optimize the parameters $\phi$ of the Octopus, the weights of LVLMs would remain frozen.

%% file: sec/5_exper.tex
\section{Experiment}
\label{sec:exper}
\subsection{Experimental Setting}

{\bf Datasets.} We conducted experiments on both generative and discriminative tasks to study the hallucinations of LVLMs. Following previous methods~\cite{woo2024don,wang2024mdpo,leng2024mitigating,favero2024multi}, we mainly build the experiments for the generative task on the AMBER~\cite{wang2023llm}, Object-HalBench~\cite{rohrbach2018object}, and MMHalBench~\cite{sun2023aligning} datasets. For the discriminative task, we evaluate the results on the AMBER~\cite{wang2023llm} and POPE~\cite{li2023evaluating} datasets. 
%

{\bf\noindent Evaluation Metric.} Following~\cite{woo2024don,rohrbach2018object,sun2023aligning}, we use four metrics to evaluate generative hallucinations on the AMBER and Object-HalBench including CHAIR~\cite{rohrbach2018object}, Cover~\cite{wang2023llm}, Hal~\cite{wang2023llm}, and Cog~\cite{wang2023llm}.
While for the MMHalBench dataset, we use GPT-4~\cite{achiam2023gpt} to evaluate the overall quality of responses.
For the discrimination task, we follow~\cite{woo2024don} and use Accuracy and F1 to measure hallucinations. 

{\bf\noindent Implementation Details.}  
To train our model, we construct two datasets for the generation and discrimination tasks. For the generation task, we build 10,000 preference data on MSCOCO~\cite{lin2014microsoft} with a language prompt ``Please Describe this image in detail.''. For the discrimination task, we follow~\cite{li2023evaluating} to build 7,000 hallucinated data from the MSCOCO~\cite{lin2014microsoft} training set.
The Adam~\cite{kingma2014adam} is used as our optimizer. We train all models on the four 3090 GPUs and the batch size is set to 4. 

\input{table/table_1}

\subsection{Quantitative Evaluation}
{\bf\noindent Generative Task.} 
In Table~\ref{tab:table1}, we show a performance comparison on the generative task, related results from ~\cite{wang2024mdpo,woo2024don}. Two general LVLMs (\emph{i.e.,} LLaVA~\cite{liu2024improved} and InstructBLIP~\cite{10.5555/3666122.3668264}) are applied to evaluate the results across three datasets including AMBER~\cite{wang2023llm}, Object-HalBench~\cite{rohrbach2018object} and MMHalBench~\cite{sun2023aligning}. In particular, we observe that our Octopus can significantly boost performance on all datasets compared with previous CD methods~\cite{leng2024mitigating,favero2024multi,woo2024don}. Meanwhile, compared to the original LLaVA model (referred to as ``Base''), our approach achieves $\sim$40\% performance improvement on the CHAIR metric of the AMBER dataset. 
Moreover, we also report the results for approaches that require retraining the entire model~\cite{sarkar2024mitigating,zhao2023beyond,yue2024less}, it can be found that Octopus still outperforms them by a large margin.

{\bf\noindent Discriminative Task.} To verify the effectiveness of our method on the discriminative task, we conduct experiments on the AMBER and POPE datasets. As shown in Tabel~\ref{tab:table2}, our model achieves significant performance gains and boosts the baseline model 9.7/11.6 and 3.75/3.02 in accuracy and F1 score for two benchmarks, respectively. 

Finally, note that the purpose of our work is not to surpass all methods across every benchmark, and we believe there remains significant room for improvement in the future by integrating more effective CD strategies.

\input{table/tabel_23}

\input{table/table_4}

\subsection{Ablation Study}

As shown in Table~\ref{tab:table4}, we conduct several ablation studies on the AMBER to demonstrate the effectiveness of our method. In the first two rows, we report the scores of the original LLaVA and corresponding results based on randomly using three CD strategies at each generative step. The results show that the randomly selected CD strategy can mitigate hallucinations to some extent. In 3-5 rows, we show the results of using two different contrastive decoding methods with our Octopus. We find that our Octopus can provide more accurate outputs compared to the random selection. Finally, when we apply three CD strategies together, the hallucination contents are eliminated to a great extent. More importantly, the above experiments demonstrate that our framework has high expansibility and the performance can be improved by introducing more ``tentacles''.




\input{table/table_5}

\subsection{Different Model Settings}
In Table ~\ref{tab:duibi}, we explore several alternative settings to further discuss the proposed method on the AMBER dataset.

{\bf\noindent Different Criterion.}
\label{duibi:kekong} 
Considering that different fields prioritize distinct metrics. For instance, the recommendation domain emphasizes recall~\cite{gunawardana2009survey}, while factual precision is more critical in the medical field~\cite{ayyamperumal2024current}.
Therefore, we focus on verifying the flexibility of our method in the first two rows. In particular, we utilize the Cover score~\cite{wang2023llm} and average score (\emph{i.e.,} averaged by Cover and CHAIR) as metrics to divide the workflow $\mathcal{A}$, respectively. 
The results show that our Octopus can be easily modified to adapt to the specific needs of different domains.   

{\bf\noindent Different Reinforcement Learning Algorithms.}
\label{duibi:rl}
In rows 3-4, we apply different RL-based algorithms to optimize our model. Specifically, we employ Monte-Carlo sampling~\cite{wang2012monte} and PPO~\cite{schulman2017proximal} to train our Octopus, respectively. 
It can be seen that these alternative reinforcement learning methods can also achieve satisfactory performance.

\begin{figure}[t]
\centering
\includegraphics[width=0.46\textwidth]{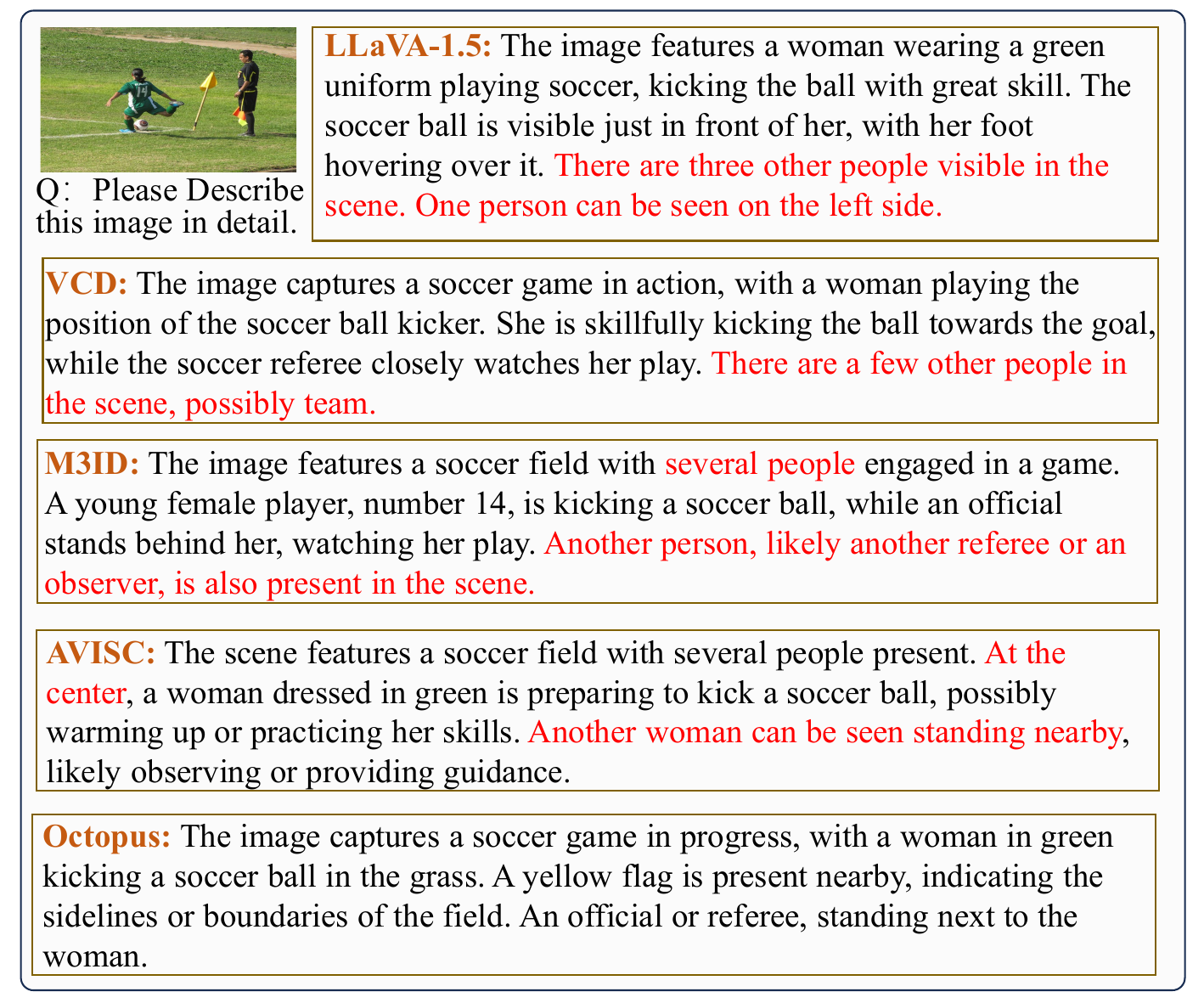}
\caption{Comparison of generated description with different CD strategies and our method. The hallucination contents are depicted by \textcolor{red}{red} text color.
}
\label{fig:fig5}
\end{figure}
{\bf\noindent Different Model Architectures.} 
In rows 5-6, we conduct experiments with different model architectures. We replace the vanilla self-attention structure with max pooling~\cite{boureau2010theoretical} and cross-attention~\cite{vaswani2017attention}, respectively. 
The results show that these transformer-based structures achieve better performance due to their enhanced modeling capabilities.
\subsection{Qualitative Evaluation.}
To further analyze and verify the proposed method, we visualize the qualitative results on AMBER dataset. As shown in Fig.~\ref{fig:fig5}, given an image and a language instruction, we present the responses of our Octopus and other methods including LLaVA-1.5~\cite{liu2024improved}, VCD~\cite{leng2024mitigating}, M3ID~\cite{favero2024multi}, AVISC~\cite{woo2024don}. The red color is used to highlight these hallucinated words.
Compared to the original outputs or single CD methods, our Octopus can better eliminate hallucinations and provide a more accurate understanding of the given image.

%% file: table/table_1.tex
\begin{table*}[t]
\centering
\small
\caption{Comparison with the state-of-the-art methods for the generative task across three datasets. 
For reference, we also provide the results of fine-tuning LVLMs. $^\dagger$ signifies results reproduced with the official implementation codes.}
\resizebox{0.95\textwidth}{!}{
\begin{tabular}{cccccccccc}\toprule
 \multirow{2}*{LVLMs}&\multirow{2}*{Method}&\multicolumn{2}{c}{MMHalBench} &\multicolumn{2}{c}{Object HalBench} &\multicolumn{4}{c}{AMBER} \\ \cmidrule(lr){3-4} \cmidrule(lr){5-6} \cmidrule(lr){7-10}
 &&Score $\uparrow$ &HalRate $\downarrow$ &CHAIR$_{s}$ $\downarrow$&CHAIR$_{i}$ $\downarrow$&CHAIR $\downarrow$&Cover. $\uparrow$ &HalRate $\downarrow$&Cog. $\downarrow$\\\midrule
  \rowcolor{lightgray} \multicolumn{10}{c}{\textbf{\textit{Referenced Results (Not Directly Comparable)}}} \\ \midrule
GPT-4V \cite{achiam2023gpt} &Base&3.49 & 0.28 &13.6 &7.3 &4.6 &67.1 &30.7 &2.6 \\ \midrule
\multirow{5}*{\shortstack{LLaVA-v1.5\\-7B\cite{liu2024improved}}}& HACL \cite{jiang2024hallucination}
&2.13 &0.50 &- &- &- &- &- &- \\
& POVID \cite{zhou2024aligning}
&2.08 &0.56 &48.1 &24.4 &- &- &- &- \\
& EOS \cite{yue2024less}
& 2.03 & 0.59 & 40.3 & 17.8 & 5.1 & 49.1 & 22.7 & 2.0 \\
& HA-DPO \cite{zhao2023beyond}
&1.97 &0.60 & 39.9 &19.9 &6.7 &49.8 &30.9 &3.3 \\
& HALVA \cite{sarkar2024mitigating}
&2.25 &0.54 &- &- &6.6 &53.0 &32.2 &3.4 \\ \midrule
 \rowcolor{lightgray} \multicolumn{10}{c}{\textbf{\textit{Comparable Results}}} \\ \midrule
 
\multirow{8}*{\shortstack{LLaVA-v1.5\\-7B \cite{liu2024improved}}}&Base
&1.59& 0.72 & 25.0  & 9.2 & 8.0  & 44.5  & 31.0  & 2.2 \\
&+ LCD \cite{manevich2024mitigating}&- &- &61.0 &16.1 &-   &- &- &- \\
 &+ ICD \cite{wang2024mitigating}
 &- &- &47.4 &13.9 &-   &- &- &- \\
&+ OPERA \cite{huang2024opera}
&2.15 & 0.54 & 45.1 & 22.3  &- &- &- &- \\
 
&+ VCD \cite{leng2024mitigating}
&1.96$^\dagger$&0.64$^\dagger$ & 23.6$^\dagger$  & 8.4$^\dagger$ & 6.7  & 46.5  & 27.8  & 2.0  \\
&+ M3ID \cite{favero2024multi}
&2.14$^\dagger$& 0.61$^\dagger$& 23.2$^\dagger$  & 7.3$^\dagger$ & 6.0  & 48.9  & 26.0  & 1.5  \\
&+ AVISC \cite{woo2024don}
&2.19$^\dagger$&0.59$^\dagger$ & 22.1$^\dagger$  & 7.8$^\dagger$ & 6.3  & 46.6  & 25.6  & 2.0  \\
&+ Octopus(Ours)  &\textbf{2.61}&\textbf{0.50} & \textbf{20.8}  & \textbf{6.6} & \textbf{4.8}  & \textbf{49.2}  & \textbf{23.4}  & \textbf{1.2}  \\ \midrule
\multirow{7}*{\shortstack{Instruct-\\BLIP \cite{10.5555/3666122.3668264}}}&Base
&1.84 &0.64 & 0.7  & 9.1 & 8.4  & 46.4  & 31.1  & 2.6 \\
&+ LCD \cite{manevich2024mitigating}&- &- &17.4 &10.7 &-   &- &- &- \\
&+ ICD \cite{wang2024mitigating}&- &- &15.2 &8.0 &-   &- &- &- \\
&+ OPERA \cite{huang2024opera}&- &- &16.6 &6.8 &-   &- &- &- \\
 
&+ VCD \cite{leng2024mitigating}
&1.75$^\dagger$ &0.64$^\dagger$ & 0.8$^\dagger$  & 8.9$^\dagger$ & 7.6  & 47.7  & 29.9  & 2.2 \\
&+ M3ID \cite{favero2024multi}
&1.70$^\dagger$ &0.65$^\dagger$ & 0.9$^\dagger$  & 7.6$^\dagger$ & 6.9  & 47.2  & 27.5  & 2.2 \\
&+ AVISC \cite{woo2024don}
&2.03$^\dagger$ &0.59$^\dagger$ & 0.7$^\dagger$ & 8.3$^\dagger$ & 6.7  & 46.7  & 28.0  & 2.6 \\
& Octopus (Ours) &\textbf{2.31} &\textbf{0.49} & \textbf{0.5}  & \textbf{6.8} & \textbf{6.1}  & \textbf{48.5}  & \textbf{22.2}  & \textbf{1.3} \\
\bottomrule
\end{tabular}
}
\label{tab:table1}
\end{table*}

%% file: table/tabel_23.tex
\begin{table*}[t]
\centering
\small
\caption{Comparison with the state-of-the-art methods for the discriminative tasks across two datasets. 
$^\dagger$ signifies results reproduced with the official implementation codes.} 
\resizebox{0.95\textwidth}{!}{
\begin{tabular}{lcccccccccc}\toprule
 &\multicolumn{2}{c}{AMBER}&\multicolumn{8}{c}{POPE\_MSCOCO} \\ 
 \cmidrule[0.4pt](lr){2-3} 
 \cmidrule[0.4pt](lr){4-11}
 &\multicolumn{2}{c}{Discrimination}&\multicolumn{2}{c}{Random}&\multicolumn{2}{c}{Popular}&\multicolumn{2}{c}{Adversarial}&\multicolumn{2}{c}{ALL}  \\
 \cmidrule[0.4pt](lr){2-3}
 \cmidrule[0.4pt](lr){4-5}
 \cmidrule[0.4pt](lr){6-7}
 \cmidrule[0.4pt](lr){8-9}
 \cmidrule[0.4pt](lr){10-11}
 &Accuracy &F1 &Accuracy &F1 &Accuracy &F1 &Accuracy &F1 &Accuracy &F1\\
 \midrule
  LLaVA-1.5-7B\cite{liu2024improved}
  & 67.00 & 71.10& 83.77  & 81.94  & 82.57  & 80.86  & 79.77  & 78.47  & 82.04  & 80.42 \\
+ICD~\cite{wang2024mitigating}
&-  &- & \textbf{87.51}& 83.28 & 83.15 & 83.91 & 79.13 & 80.41&83.26&82.53   \\
  +ConVis~\cite{park2024convis}& - &- &84.70  &-   &83.20   &-   &81.10  &-  &83.00  &-   \\
  +OPERA~\cite{huang2024opera}& - &- &84.40  &-   &83.40   &-   &81.20  &-  &83.00  &-   \\
 +VCD~\cite{leng2024mitigating}
 & 67.30 & 71.10& 85.43  & 83.99  & 83.17  & 81.94  & 80.27 & 79.49  & 82.96  & 81.81 \\
 +M3ID~\cite{favero2024multi}
 $^\dagger$
 & 67.25& 70.90& 86.13  & 81.85  & 82.07  & 80.77  & 79.50  & 78.15  & 82.57  & 80.26  \\
  +AVISC~\cite{woo2024don}
  & 70.70& 75.45& 84.67  & 82.21  & 83.67  & 81.27  & 81.83  & 79.55  & 83.39  & 81.01  \\
 +Octopus (Ours)& \textbf{76.70} & \textbf{82.70}& \textbf{87.51}  & \textbf{85.40}  & \textbf{85.20}  & \textbf{84.19}  & \textbf{82.22}  & \textbf{81.44}  & \textbf{85.79}  & \textbf{83.44} \\
  \midrule
  InstructBLIP~\cite{10.5555/3666122.3668264}
  & 68.20& 74.60& 81.53  & 81.19  & 78.47  & 78.75  & 77.43  & 78.00  & 79.14  & 79.31 \\
  +ICD~\cite{wang2024mitigating}
  &-&-&84.36& 83.82& 77.88& 78.70 & 75.17 & 77.23 &79.14  &79.92    \\
  +OPERA~\cite{huang2024opera}
  & - & -&84.57& 83.74 & 78.24& 79.15 & 74.59 & 76.33 &79.13  &79.74   \\
  +VCD~\cite{leng2024mitigating}
  & 69.65& 75.90& 82.03  & 81.56  & 79.13  & 79.20  & 77.23  & 77.72  & 79.46  & 79.49 \\
  +M3ID~\cite{favero2024multi}
  $^\dagger$
  & 69.05& 75.25& 82.33  & 81.53  & 80.90  & 80.42  & 78.53  & 78.49  & 80.59  & 80.15 \\
  +AVISC~\cite{woo2024don}
  & 72.60& 78.60& 86.03  & 84.41  & 84.27  & 82.77  & 81.83  & 80.67  & 84.04  & 82.62 \\
  +Octopus (Ours)& \textbf{74.00}& \textbf{79.70}& \textbf{86.63}  & \textbf{85.30}  & \textbf{84.90}  & \textbf{83.55}  & \textbf{82.83}  & \textbf{81.43}  & \textbf{84.79}  & \textbf{83.43}  \\

\bottomrule
\end{tabular}
}
\label{tab:table2}
\end{table*}

%% file: table/table_4.tex
\begin{table}[t]
\centering
\small
\caption{\textbf{Ablation study.} We select different numbers of contrastive decoding methods as candidates to demonstrate the effectiveness of our approach. ``Str1, Str2, Str3'' indicate CD strategy VCD~\cite{leng2024mitigating}, M3ID~\cite{favero2024multi} and AVISC~\cite{woo2024don} in Sec.~\ref{sec:sec3.2}, respectively.}
\resizebox{0.48\textwidth}{!}{
\begin{tabular}{ccccccccc}\toprule
 &Str1&Str2&Str3& Octopus &CHAIR $\downarrow$&Cover. $\uparrow$ &Hal. $\downarrow$&Cog. $\downarrow$\\
 \midrule
  1&&&& & 8.0  & 44.5  & 31.0  & 2.2  \\
 2 &\checkmark &\checkmark &\checkmark& &6.9&46.2&26.1&2.2\\
  \midrule
 3&\checkmark &\checkmark & &\checkmark & 5.5  & 48.7  & 25.8  & 1.5 \\
 4& \checkmark && \checkmark& \checkmark &5.7  & 48.2  & 25.3  & 1.5 \\
 5 & &\checkmark &\checkmark&\checkmark & 5.5  & 48.4  & 26.2  & 1.6 \\
 6 &\checkmark &\checkmark &\checkmark &\checkmark & \textbf{4.8}  & \textbf{49.2} & \textbf{23.4}  & \textbf{1.2} \\

\bottomrule
\end{tabular}
}

\label{tab:table4}
\end{table}

%% file: table/table_5.tex
\begin{table}[t]
\centering
\small
\caption{The effects of different settings for our method.}
\label{tab:table5}
\resizebox{0.42\textwidth}{!}{
\begin{tabular}{cccccc}\toprule

  &&CHAIR $\downarrow$&Cover. $\uparrow$ &Hal. $\downarrow$&Cog. $\downarrow$ \\
  \midrule
  1&Cover&5.4&\textbf{50.1}&26.1&1.2 \\
  2&Average&5.0&49.8&23.0&1.4\\
  \midrule
  3&Monte Carlo~\cite{wang2012monte}
&5.1&48.1&23.9&1.3\\
  4&PPO~\cite{schulman2017proximal}&5.8&47.5&\textbf{23.0}&1.5 \\
  \midrule
  5&Pooling~\cite{boureau2010theoretical}&6.2&44.8&25.0&1.6 \\
  6&Cross-atteition~\cite{vaswani2017attention}&5.1&49.0&24.2&1.3 \\
  \midrule
  7&Ours&\textbf{4.8}  & 49.2  & 23.4  & \textbf{1.2} \\

\bottomrule
\end{tabular}
}
\label{tab:duibi}
\end{table}

%% file: sec/6_conclusion.tex
\section{Conclusion}
\label{sec:conclu}
In this paper, we first explore the mechanism behind hallucination occurrences. Extensive experiments show that hallucination causes are hybrid and each generative step suffers from different challenges. 
Based on the above insights, we propose a new framework Octopus that can adaptively classify hallucination types and build dynamic workflows for different inputs. More importantly, the Octopus has excellent deployability and expansibility, making it a versatile tool for various fields. We expect that this work can provide a general framework to alleviate hallucination challenges across different scenarios.